# MESD: EXPLORING OPTICAL FLOW ASSESSMENT ON EDGE OF MOTION OBJECTS WITH MOTION EDGE STRUCTURE DIFFERENCE

*Bin Liao[1], Jinlong Hu[2]*

[1] College of Mathematics and Informatics, South China Agricultural University, Guangzhou, China
[2] Guangdong Key Lab of Communication and Computer Network, School of Computer Science and Engineering, South China University of Technology, Guangzhou, China
liaobin_lb@scau.edu.cn, jlhu@scut.edu.cn

**ABSTRACT**

The optical flow estimation has been assessed in various applications. In this paper, we propose a novel method named motion edge structure difference(MESD) to assess estimation errors of optical flow fields on edge of motion objects. We implement comparison experiments for MESD by evaluating five representative optical flow algorithms on four popular benchmarks: MPI Sintel, Middlebury, KITTI 2012 and KITTI 2015. Our experimental results demonstrate that MESD can reasonably and discriminatively assess estimation errors of optical flow fields on motion edge. The results indicate that MESD could be a supplementary metric to existing general assessment metrics for evaluating optical flow algorithms in related computer vision applications.

## 1. INTRODUCTION

Optical flow estimation has been widely used in various applications, such as surveillance scene analysis [1], action recognition [2] and flying robots control [3]. The average end-point-error(AEPE) [4] is a major assessment metric for optical flow estimation in recent state-of-the-art works such as RAFT [5], PWC-Net_ROB [6], LiteFlowNet [7] and FlowNet2 [8]. The improvement of optical flow algorithms is usually expressed as the decrease of AEPE, but the AEPE is insufficient to describe estimation errors of optical flow fields in some regions related to motion recognition, motion objects segmentation and motion understanding. Dosovitskiy et al. [9] present that FlowNetS often better preserves fine details than EpicFlow [10] does, although the AEPE of FlowNetS is usually worse than EpicFlow. Wang et al.[11] achieve good performance on four public action recognition datasets by adopting LDOF [12]. The comparison experiments in action recognition reveal that LDOF produces the best performance comparing to different optical flow methods although LDOF yields large AEPE [13]. The AEPE is not well correlated to the performance of action recognition, motion object detection and objects segmentation [13][14]. Further experiments reveal that some difference among various optical flow methods is at the boundary of motion objects and inside the motion objects [13][15][16]. Accuracy of optical flow on motion boundaries and small displacements is closely related to the performance of action recognition and motion objects segmentation. Different assessment metrics of optical flow are useful for specific applications [17].

The study of visuomotor perception reveals that the motion edge provides an important cue for seeing object outline when an object moves against its background [18][19]. Pasupathy et al.'s experimental results[20] suggest that shape selectivity in most V4 neurons likely arises by pooling both surface contrast and boundary contour information, and such a strategy would facilitate the segmentation of objects from natural scenes.

Inspired by the role of perceptual contours in visuomotor perception and boundary mechanisms possessed by humans for detecting motion boundaries, we propose an assessment metric named motion edge structure difference (MESD) to assess estimation errors of optical flow fields on motion edge in this paper. The experimental results demonstrate that MESD can reasonably and discriminatively measure optical flow fields errors on motion edge.

## 2. METHODS

Edge of optical flow fields usually represents motion edges. Human beings can distinguish motion edge by extracting structural information [21] from optical flow fields. Cognitive vision studies [22] have shown that human vision system is highly sensitive to structures such as global information and local details in scenes. The proposed

assessment metric would take into account human visuomotor perception:

- consider the region-level structure similarity of motion edge and the object-level properties of motion details.
- be insensitive to slight mismatches of background.
- be capable of capturing holistic content.
- closely match human perception so that good motion objects edge can be directly used in various computer vision applications such as action recognition and motion objects segmentation.

Human psychophysical experiments suggest that neurons [18] only encode contours that are associated with a local contrast in texture and luminance across the boundary. Contours of motion objects usually correspond to the area with obvious contrast in the optical flow fields. To assess difference of motion edge between ground truth and estimated optical flow, we firstly compute gradient on optical flow fields by using horizontal and vertical filter templates on the ground truth and the estimated optical flow field as follows:

$$g_x = \frac{1}{2}[1 \quad -1] \qquad g_y = \frac{1}{2}\begin{bmatrix} 1 \\ -1 \end{bmatrix} \qquad (1)$$

Where $x$ and $y$ represent horizontal and vertical directions, respectively.

$u(x,y)$ and $v(x,y)$ are a horizontal and a vertical component of optical flow. $u_o$ and $v_o$ represent the horizontal and vertical component of ground truth, respectively. $u_f$ and $v_f$ are the horizontal and vertical component of the estimated optical flow field. $u_{ox}$, $u_{oy}$, $v_{ox}$ and $v_{oy}$ are the horizontal and vertical gradient of ground truth as follows:

$$u_{ox} = u_o \otimes g_x, u_{oy} = u_o \otimes g_y, v_{ox} = v_o \otimes g_x, v_{oy} = v_o \otimes g_y$$
$$u_{fx} = u_f \otimes g_x, u_{fy} = u_f \otimes g_y, v_{fx} = v_f \otimes g_x, v_{fy} = v_f \otimes g_y \quad (2)$$

Where $u_{fx}$, $u_{fy}$, $v_{fx}$ and $v_{fy}$ are the horizontal and vertical gradient of the estimated optical flow.

The edge structure similarity (ESS) is formulated by the product of luminance comparison, contrast comparison and structure comparison of gradient.

$$ESS(u_{ox}, u_{fx}) = \frac{2\mu(u_{ox})\mu(u_{fx})}{(\mu(u_{ox}))^2 + (\mu(u_{fx}))^2} \cdot \frac{2\sigma(u_{ox})\sigma(u_{fx})}{(\sigma(u_{ox}))^2 + (\sigma(u_{fx}))^2} \cdot \frac{\sigma(u_{ox}, u_{fx})}{\sigma(u_{ox})\sigma(u_{fx})} \quad (3)$$

Where $\mu(u_{ox})$ and $\mu(u_{fx})$ are the mean of $u_{ox}$ and $u_{fx}$, respectively. $\sigma(u_{ox})$ and $\sigma(u_{fx})$ are the standard deviation of $u_{ox}$ and $u_{fx}$, respectively. $\sigma(u_{ox}, u_{fx})$ is the covariance of $u_{ox}$ and $u_{fx}$.

The MESD is expressed as the motion edge structure difference between ground truth and estimated optical flow as follows:

$$MESD = \left(1 - \frac{1}{4} \cdot \left(ESS(u_{ox}, u_{fx}) + ESS(u_{oy}, u_{fy})\right.\right. \quad (4)$$
$$\left.\left. + ESS(v_{ox}, v_{fx}) + ESS(v_{oy}, v_{fy})\right)\right) \times 100\%$$

## 3. EXPERIMENTS

To explore the effectiveness of MESD and the relationship between MESD and AEPE, we utilized MESD and AEPE to evaluate five representative optical flow algorithms and their modified methods on four popular benchmarks.

### 3.1. Benchmarks

There are different characteristics on MPI Sintel [23], Middlebury [24], KITTI 2012 and KITTI 2015 [25]. MPI Sintel derives from the open source 3D animated short film including different types motion. Most of data in Middlebury datasets is about indoor small displacements. KITTI focuses on autonomous driving for vehicles with non-100% density ground truth.

### 3.2. Optical flow algorithms for evaluation

We use MESD and AEPE to evaluate five representative methods. (i) MDP-Flow2 [26], a representative variational framework; (ii) LDOF [12], good performance in action recognition; (iii) FlowNet2 [8], the successor of first training end-to-end CNNs for optical flow; (iv) PWC-Net_ROB [6], a popular optical flow method based on CNNs; (v) LiteFlowNet [7], training end-to-end CNNs with sub-pixel refinement units. M, L, F, P and Li represent MDP-Flow2, LDOF, Flownet2, PWC-Net_ROB and Liteflownet, respectively.

### 3.3. Edge refinement

To yield different accuracy on motion edge for the same method in comparison experiments, we adopted edge refinement (ER) as an optical flow post-processing. The weight $w_{i,j}(i',j')$ of pixel $(i',j')$ is determined by (5) according to their spatial distance and colour distance.

$$w_{i,j}(i',j') = \exp\left\{-0.5 \times \left[\frac{|i-i'|^2 + |j-j'|^2}{n_1^2} + \frac{|I(i,j) - I(i',j')|^2}{n_2^2 C}\right]\right\} \quad (5)$$

Where $(i',j')$ is a pixel in the neighbourhood $N_{i,j}$ of pixel $(i,j)$. $I$ is a colour frame, and $C$ is the number of colour channels. We set $n_1$ and $n_2$ equal to 7.

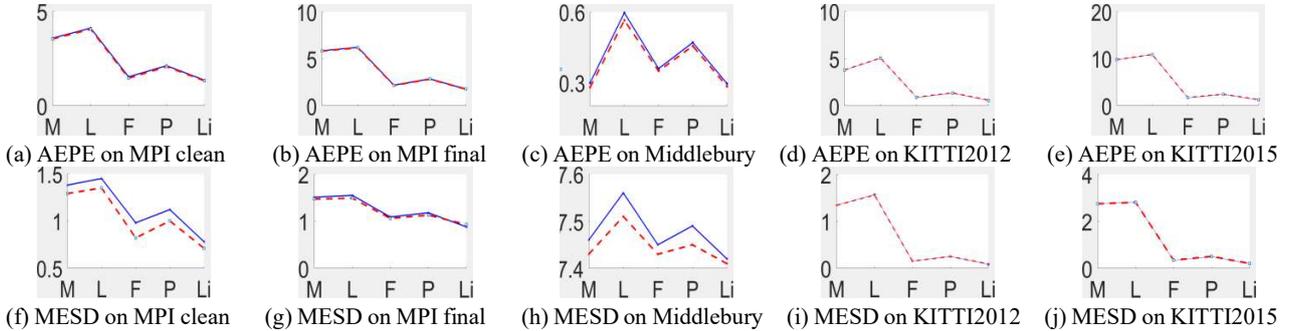

**Fig.1.** Comparison of AEPE and MESD. AEPE and MESD are shown in first and second row, respectively. The training datasets results of MPI-Sintel clean, MPI-Sintel final, Middlebury, KITTI2012 and KITTI2015 are illustrated from left to right column. M, L, F, P and Li represent MDP-Flow2, LDOF, Flownet2, PWC-Net_ROB and Liteflownet, respectively. Blue dot(solid line) and red dot(dashed line) represent results of five optical flow algorithms and their modified methods using ER.

The post-processing optical flow $UV(i,j)$ on pixel $(i,j)$ is determined by (6).

$$UV(i,j) = median\{w_{i,j}(i',j') * uv(i',j')\}_{N_{i,j}} \quad (6)$$

Where $uv(i',j')$ is the optical flow on pixel $(i',j')$ before post-processing. We utilized (6) as a post-processing on the five algorithms to obtain different accuracy on motion edge.

### 3.4. Experimental procedure

We firstly estimated optical flow by five algorithms. Then we obtained different optical flow accuracy on motion edge by using ER as a post-processing. We compared the difference of MESD between the results of original methods and ER. We also compared AEPE and MESD on popular benchmarks to explore the relationship between AEPE and MESD.

### 4. RESULTS AND DISCUSSION

Fig. 1 shows the MESD and AEPE of five optical flow algorithms and their modified methods with ER. Fig. 2 illustrates an example having the same change trend of AEPE and MESD on large displacements. Fig. 3 and Fig. 4 show examples having the different change trend of AEPE and MESD on small displacements and motion edge.

### 4.1. The discrimination ability of MESD

The AEPE and MESD of comparison methods on four popular benchmarks are presented in Fig. 1. We used valid pixels of ground truth to compute MESD on KITTI datasets (NOC). It shows that AEPE and MESD have the same change trends on the four public benchmarks in Fig. 1. The ER slightly reduces the AEPE, and MESD achieves more obvious relative improvement especially on MPI-Sintel training datasets. Assessed by MESD, Flownet2 using ER(Flownet2-ER) achieved about 16.3% relative improvement on MPI-Sintel training clean dataset. It can be observed that the ER described in (6) as a post-processing reduces both AEPE and MESD of all five comparison methods from Fig. 1. The relative improvement assessed by MESD is more obvious as shown in Fig. 1f and Fig. 1h, and the reason for this may be that the ER mainly enhances accuracy of motion edge. It indicates that MESD can reasonably assess accuracy improvement of optical flow in general as shown in Fig. 1, and the ER as a post-processing is helpful to improve the accuracy of optical flow.

Fig. 2 presents a comparison example on large displacements. As shown in Fig. 2a, Fig. 2b and Fig. 2c, PWC-Net_ROB-ER yields 3.05% and 20% relative improvement of AEPE and MESD on the whole image comparing with PWC-Net_ROB. The PWC-Net_ROB-ER achieves 6.08% and 17.21% relative improvement of AEPE and MESD in the main motion foreground as shown in Fig. 2d, Fig. 2e and Fig. 2f. The PWC-Net_ROB-ER yields more accurate motion edge as displayed in Fig. 2g, Fig. 2h and Fig. 2i. Such characteristic of MESD indicates that MESD is suitable to describe the accuracy of optical flow especially on motion edge.

### 4.2. Difference between AEPE and MESD

On the four comparison benchmarks, there are the same order ranked by MESD and AEPE of comparison methods and their ER methods as shown in Fig. 1 in general. While we noticed that the change trends of MESD are different from AEPE in some cases.

Fig. 3 presents a comparison example on small displacements motion. Fig. 3a and Fig. 3b are the Sintel training clean sleeping_2 frame0005 and frame0006, respectively. There is a lot of small displacements motion as displayed in Fig. 3c and Fig. 3d. From Fig. 3e to Fig. 3i, MDP-Flow2-ER achieves the least AEPE among the five methods. MDP-Flow2-ER and Liteflownet-ER yield the least MESD as shown in Fig. 3j and Fig. 3n. The reason for this different change trend between AEPE and MESD in Fig. 3 may be that AEPE computes optical flow errors on each pixel, while the MESD focuses on assessing accuracy of optical flow on motion edge without considering motion region inner. The difference between AEPE and MESD is obvious when there are a lot of small displacements motion inside the motion region.

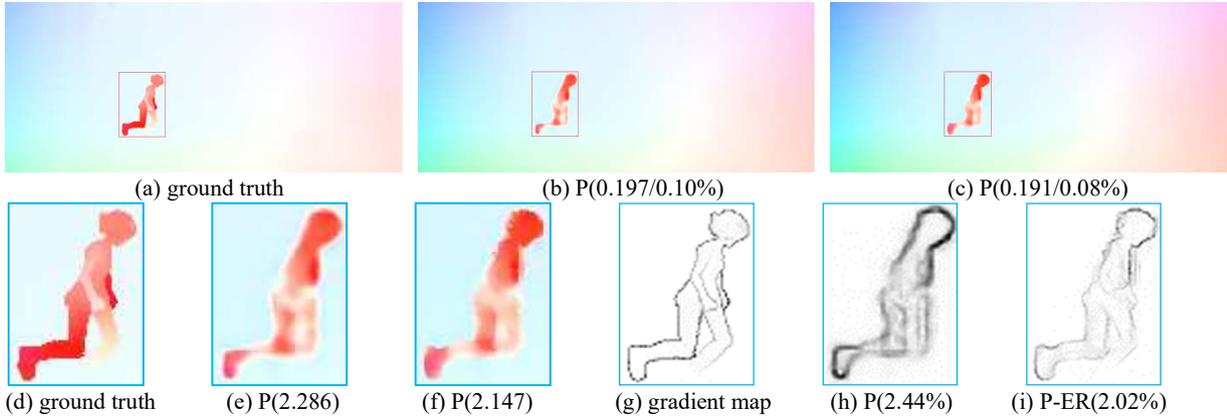

**Fig. 2.** The AEPE and MESD on Sintel training clean alley_2 frame_0005. (a), (b), (c) are the colour code images of ground truth, PWC-Net_ROB and PWC-Net_ROB-ER, respectively. AEPE/MESD is presented for each colour code image, for example 0.197/0.1%. Main motion objects labelled with red box are amplified and shown in (d), (e), (f), and their gradient maps are displayed in (g), (h), (i). 2.286 and 2.147 are AEPE of (e) and (f). 2.44% and 2.02% are MESD of (h) and (i).

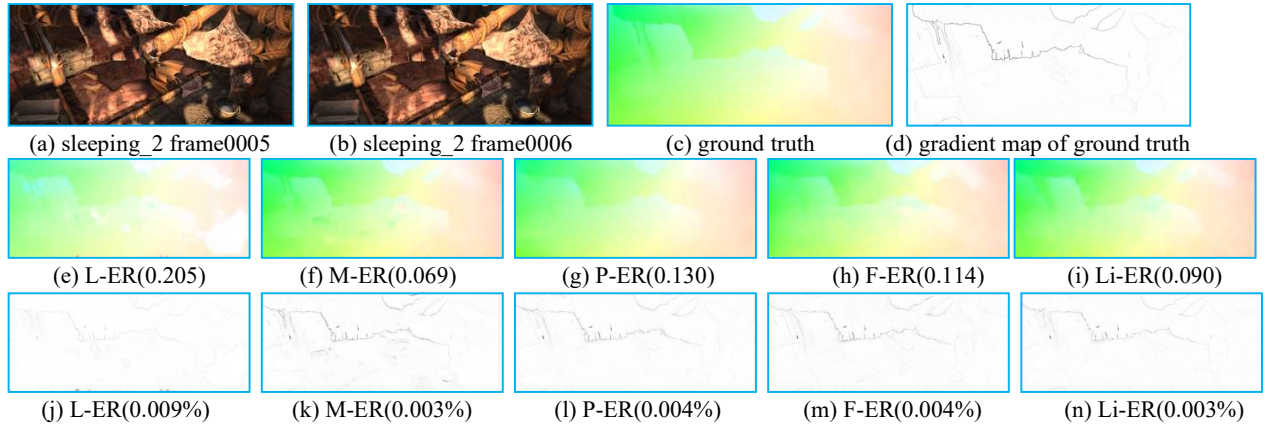

**Fig. 3.** The results of small displacements. (a) and (b) are the two frames of Sintel training clean sequences. (c) and (d) are the color coding image and the gradient map of ground truth. Optical flow color coding images and AEPE of five methods are displayed from (e) to (i). Gradient difference maps labelled with MESD are shown from (j) to (n), and darker black means larger gradient difference.

Fig. 4 presents an example on large displacements motion. The motion foreground has more large displacements motion than background does as shown in Fig. 4a. Flownet2-ER has 0.09% relative increase assessed by AEPE on the whole frame as displayed in Fig. 4b, while there is 1.56% relative decrease assessed by MESD as shown in Fig. 4c.

The motion foreground objects are illustrated from Fig. 4d to Fig. 4o. The AEPE on motion objects are 3.264 and 3.120 and the MESD of these objects are 6.18% and 5.96% as shown from Fig. 4d to Fig. 4i. The relative improvement of Flownet2-ER is 4.41% on AEPE and 3.56% on MESD as shown from Fig. 4d to Fig. 4i. The FlowNet2-ER also yields less AEPE and MESD than FlowNet2 does on the other motion foreground objects as shown from Fig. 4j to Fig. 4o. As shown from Fig. 4d to Fig. 4o, the FlowNet2-ER obtains less AEPE and MESD on all motion foreground objects than Flownet2 does. The change trend of AEPE on all motion foreground objects is different from the change trend of AEPE on the whole frame, while the change trend of MESD on all motion foreground objects is consistent with the change trend of MESD on the whole frame as illustrated in Fig. 4.

As shown in Fig. 4a, the region occupied by motion foreground objects is small and background has obvious motion. In such case, background errors are prone to become the main indicator of errors assessed by AEPE because the AEPE is calculated on each pixel. Optical flow accuracy improvement of small objects on some special region such as motion edge cannot reduce the AEPE of the whole frame as shown in Fig. 4, while MESD can indicate the accuracy improvement on motion edge of small objects as displayed in Fig. 4. Such inconsistent change trend demonstrates that the MESD and AEPE assess the accuracy of optical flow from different aspects, and MESD focuses on quantitatively assessing optical flow accuracy on motion object edge.

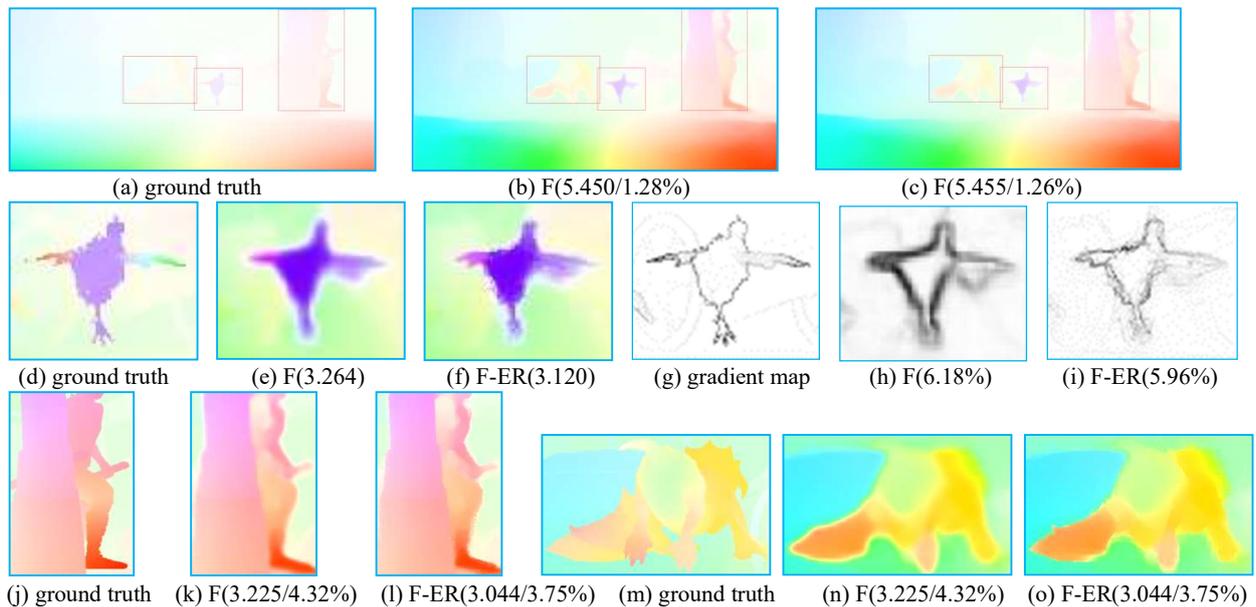

**Fig. 4.** An example of the different change trend between AEPE and MESD. (a), (b), (c) are optical flow colour code images of ground truth(training clean market_6 frame0006), Flownet2 and Flownet2-ER. AEPE/MESD is presented in (b), (c), (k), (l), (n) and (o), such as 5.450/1.28%. Main motion objects labelled with red box are amplified and shown from (d) to (o). The (g), (h) and (i) are the gradient maps of (d), (e) and (f), respectively. AEPE are shown in (e) and (f) (3.264 and 3.120). The MESD of the same motion object are 6.18% and 5.96% as shown in (h) and (i).

## 5. CONCLUSION

Evaluating optical flow estimation on motion boundary is helpful to select appropriate optical flow algorithms for applications related to motion boundary, such as action recognition, motion objects segmentation and motion analysis. In this paper we propose an optical flow assessment metric named MESD inspired by boundary mechanisms possessed by humans for visuomotor perception. The MESD focuses on evaluating optical flow methods on motion edge. Experimental results demonstrate that the MESD can discriminatively and quantitatively measure the optical flow errors on motion edge and details. More comprehensive understanding of optical flow algorithms could be acquired by using MESD and AEPE to evaluate optical flow methods.

The MESD could be a supplementary metric to evaluate optical flow estimation in various computer vision applications such as action recognition, action understanding and motion analysis.